\documentclass[10pt,twocolumn,letterpaper]{article}

\usepackage[pagenumbers]{cvpr} 

\usepackage{graphicx}
\usepackage{amsmath}
\usepackage{amssymb}
\usepackage{booktabs}
\usepackage{soul}
\usepackage{subfiles} 
\usepackage{caption}
\usepackage{subcaption}
\usepackage{multirow}
\usepackage{makecell}
\usepackage{array}
\usepackage{colortbl} 
\usepackage{lipsum}
\usepackage{listings}
\usepackage{enumitem}
\usepackage[toc,page]{appendix}

\newcolumntype{+}{>{\global\let\currentrowstyle\relax}}
\newcolumntype{^}{>{\currentrowstyle}}
\newcolumntype{s}{>{\columncolor{black!10}} r}

\usepackage[dvipsnames]{xcolor}
\definecolor{C0}{rgb}{0.26,0.52,0.95}
\definecolor{C1}{rgb}{0.98,0.73,0.02}
\definecolor{C2}{rgb}{1.00,0.60,0.00}
\definecolor{C3}{rgb}{0.467,0.467,0.467}
\definecolor{C4}{rgb}{0.984,0.757,0.369}
\definecolor{C5}{rgb}{0.557,0.729,0.259}
\definecolor{C6}{rgb}{1.0,0.71,0.722}

\makeatletter
\@namedef{ver@everyshi.sty}{}
\makeatother
\usepackage{tikz}
\usetikzlibrary{positioning}
\usetikzlibrary{backgrounds}

\usepackage[pagebackref,breaklinks,colorlinks]{hyperref}

\usepackage[capitalize]{cleveref}
\crefname{section}{Sec.}{Secs.}
\Crefname{section}{Section}{Sections}
\Crefname{table}{Table}{Tables}
\crefname{table}{Tab.}{Tabs.}

\def\cvprPaperID{12879} 
\def\confName{CVPR}
\def\confYear{2024}

\newcommand{\ours}{Modeling Collaborator\xspace}
\newcommand{\supplementary}{Appendix\xspace}

\begin{document}

\title{\ours: Enabling Subjective Vision Classification With Minimal Human Effort via LLM Tool-Use}

\author{
Imad Eddine Toubal$^{1,2}$\thanks{This work was done during an internship at Google.} \hspace{2mm}
Aditya Avinash$^{1}$ \hspace{2mm}
Neil Gordon Alldrin$^{1}$ \hspace{2mm}
Jan Dlabal$^{1}$ \hspace{2mm}
Wenlei Zhou$^{1}$ \\ \hspace{2mm}
Enming Luo$^{1}$ \hspace{2mm}
Otilia Stretcu$^{1}$ \hspace{2mm}
Hao Xiong$^{1}$ \hspace{2mm}
Chun-Ta Lu$^{1}$ \hspace{2mm}
Howard Zhou$^{1}$  \\ \hspace{2mm}
Ranjay Krishna$^{1,3}$ \thanks{This work was done during working at Google.} \hspace{2mm} \vspace{2mm}
Ariel Fuxman$^{1}$ \hspace{2mm}
Tom Duerig$^{1}$ \\ \hspace{2mm}
$^1$Google Research \hspace{3mm} $^2$University of Missouri \hspace{2mm} $^3$University of Washington \\
{
\small \tt{itdfh@umsystem.edu, \{adity,nalldrin\}@google.com}
}
}

\maketitle

\begin{abstract}

From content moderation to wildlife conservation, the number of applications that require models to recognize nuanced or subjective visual concepts is growing. Traditionally, developing classifiers for such concepts requires substantial manual effort measured in hours, days, or even months to identify and annotate data needed for training. Even with recently proposed Agile Modeling techniques, which enable rapid bootstrapping of image classifiers, users are still required to spend 30 minutes or more of monotonous, repetitive data labeling just to train a single classifier. Drawing on Fiske’s Cognitive Miser theory, we propose a new framework that alleviates manual effort by replacing human labeling with natural language interactions, reducing the total effort required to define a concept by an order of magnitude: from labeling 2,000 images to only 100 plus some natural language interactions. Our framework leverages recent advances in foundation models, both large language models and vision-language models, to carve out the concept space through conversation and by automatically labeling training data points. Most importantly, our framework eliminates the need for crowd-sourced annotations. Moreover, our framework ultimately produces lightweight classification models that are deployable in cost-sensitive scenarios. Across 15 subjective concepts and across 2 public image classification datasets, our trained models outperform traditional Agile Modeling as well as state-of-the-art zero-shot classification models like ALIGN, CLIP, CuPL, and large visual question answering models like PaLI-X. 

\end{abstract}

\vspace{-0.5cm}
\section{Introduction} 
\label{sec:intro}
\begin{figure}[ht]
    \centering
    \includegraphics[width=\linewidth]{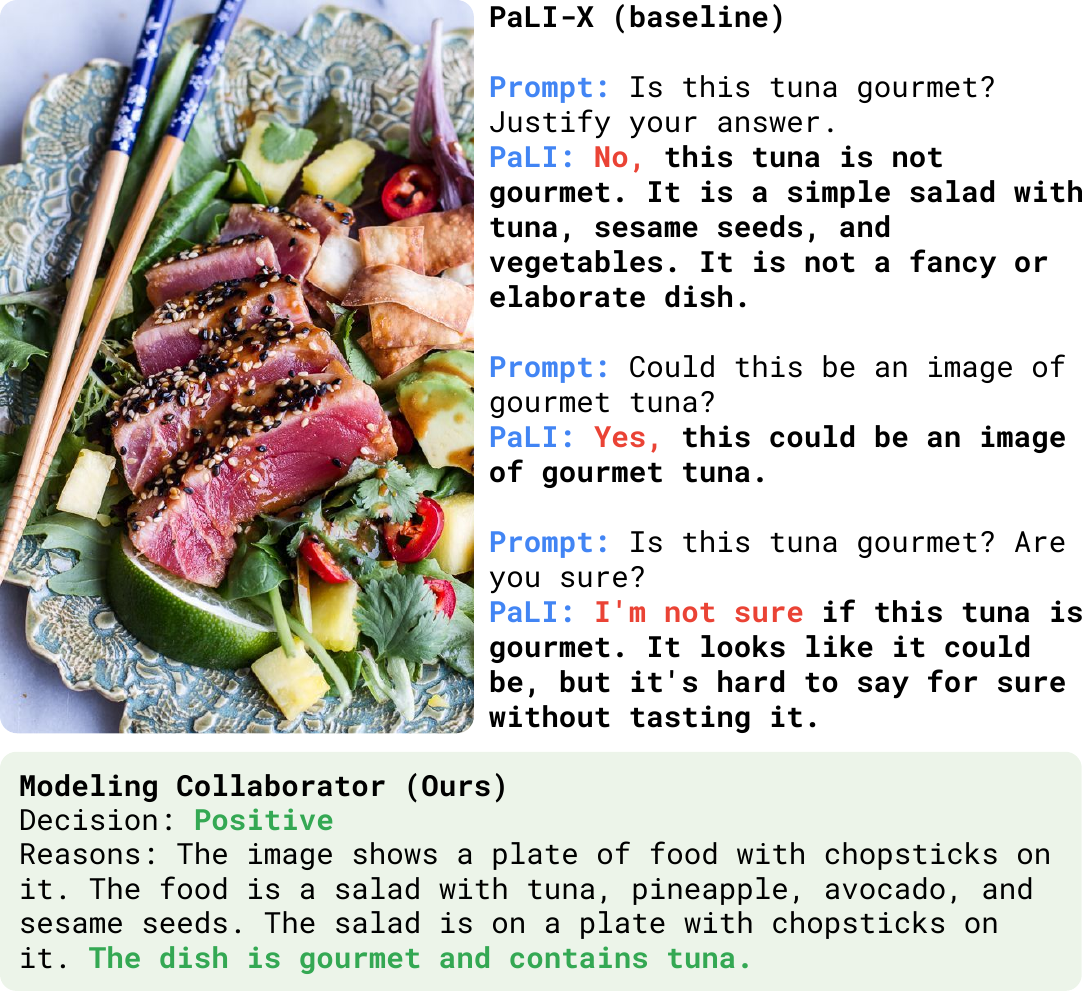}
    \caption{We introduce \ours: a framework that allows anyone to train vision models using natural language interactions and minimal effort. We show that today's best models (e.g.~PaLI-X~\cite{chen2023pali}) change their answers depending on the prompt when classifying subjective concepts like \texttt{gourmet tuna}. Meanwhile, \ours~ uses LLMs and tool-use to train vision models by interacting with users to carve out the concept space.}
    \label{fig:pali-vs-modeling-copilot}
\end{figure}

The field of computer vision has primarily focused on recognizing concepts that are objectively agreed upon, such as dogs, cats, or cars~\cite{deng2009imagenet,lin2014microsoft,kuznetsova2020open}. Even research on fine-grained recognition (e.g. ``black footed albatross'') and compositional concepts (e.g., ``red car next to a motorcycle'') have universal consensus~\cite{krishna2017visual,ji2020action,lu2016visual,ma2023crepe}. 
However, many practical real-world vision applications frequently involve recognizing subjective concepts that suffer from significant disagreements amongst individuals. Applications include predicting emotions, measuring aesthetic appeal, or content moderation~\cite{khosla2015understanding,connie2017smart,roy2017automated,kiela2020hateful}. A content moderator needs a model to identify unsafe content according to their definition of what constitutes as \textit{unsafe}; a food critic might not consider a tuna sandwich to be gourmet while others might (Figure~\ref{fig:pali-vs-modeling-copilot}).
To operationalize these applications, we need user-centric training frameworks that enable anyone to train subjective vision models.

Recently, Agile Modeling formalized the process for turning any visual concept into a vision model through a user-in-the-loop framework~\cite{stretcu2023agile}. Their work concluded that crowd workers struggled to produce labels that were consistent with the user's concept definition. Instead, they proposed an active learning algorithm, where the user iteratively labels a series of training images themselves. Unfortunately, this process is tedious, repetitive, and labor intensive; users had to label $\sim2000$ images, which on average took 30 minutes to train a binary classifier.

Existing processes fall short because they do not leverage a key capability that humans possess.
People are adept at breaking down complex subjective concepts into more manageable and objective components by applying first-order logic~\cite{miller1956magical,frege1879begriffsschrift}. This ability can be explained using Susan Fiske's Cognitive Miser Theory: people decompose complex work to avoid high cognitive load~\cite{fiske1991social}. 
People apply the same process to define complex concepts such as ``unsafe'' and ``gourmet''.
For instance, one food critic might decompose the subjective concept of ``gourmet'' as images that need to at least contain ``tuna''; if it is ``ahi tuna'', then it is likely gourmet; if it is ``canned'', then it is unlikely to be gourmet; if the dish is a ``sandwich'', then it is still not gourmet.
This decomposition of the subject concept ``gourmet'' into conjunction clauses of objective concepts ``ahi tuna'', ``canned'', and ``sandwich'' is a simple \textit{non}-laborious, cognitively effortless conversion.

With this grounding, we deliver \textbf{\ours} which empowers users to build classifiers while minimizing manual effort. 
Instead of asking users to annotate thousands of images~\cite{stretcu2023agile}, \ours~ requires $100$, along with a few natural language interactions that help decompose subjective concepts into its constituent sub-components.
To enable \ours, we leverage advancements in large language models (LLMs)~\cite{brown2020gpt3,devlin2018bert,openai2023gpt4,chowdhery2022palm,anil2023palm2} and in particular, their ability to use vision-language models (VLMs)~\cite{chen2023pali,chen2023pali3,chen2022pali} and other tools~\cite{hsieh2023tool}.
When users have a concept in mind and use \ours, it employs an LLM, which breaks the concept into questions that are digestible for a Visual Question Answering (VQA) model~\cite{chen2022pali}.
The LLM then summarizes the answers provided by the VQA model and performs reasoning through chain-of-thought~\cite{wei2022chain} to classify new images as positive or negative examples of the concept.
Users are only asked to manually label a small $100$ image validation set.
Finally, \ours labels a large amount of unlabeled images available online and uses it as distillation data to train a light-weight deployment-ready vision model.

Our method is shown to outperform existing zero-shot methods (CLIP \cite{radford2021learning}, CuPL\cite{pratt2022does} and PaLI-X\cite{chen2023pali}), especially on harder subjective concepts. When compared to the original Agile Modeling~\cite{stretcu2023agile} our system exceeds the quality of crowd-raters on hard concepts while simultaneously reducing the need for manual user-provided ground-truth by orders of magnitude.
By reducing the barriers of manual effort and resulting costs needed to develop classification models, it will empower users to rapidly convert their ideas into reality. This, in turn, has the potential to usher in a new wave of end-user applications.
\begin{figure*}[ht]
    \centering
    \includegraphics[width=\linewidth]{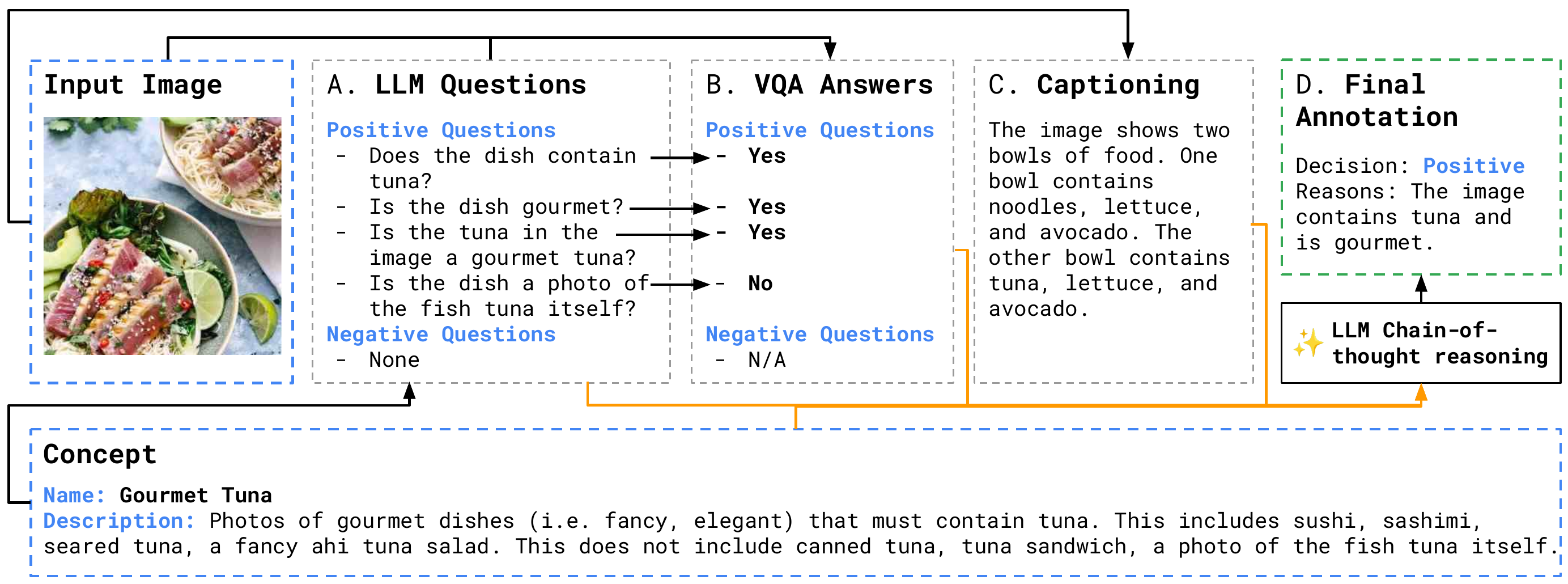}
    \caption{Modeling Collaborator Annotator system. For a given image, concept name, and description, the Annotator outputs a positive or negative label. Based on the name and description of the concept, the LLM generates relevant atomic questions to ask a VQA model (PaLI VQA in our case) (step A). These questions are fed into the VQA model that typically outputs a yes/no short answer (Step B). Additionally, we use a captioning version of PaLI (Step C) to generate a detailed description capturing as much detail as possible from the image. Finally, the LLM goes through a chain-of-thought reasoning process to output a decision and rationale (Step D).}
    \label{fig:methods-annotation-process}
\end{figure*}
\section{Related work} 
\label{sec:related}

Our work draws on advances in VLMs and LLMs and provides an improved solution to the recently introduced Agile Modeling problem.

\noindent\textbf{Agile Modeling.}
Inspired by agile software development, Agile Modeling \cite{stretcu2023agile} focuses on rapid development of image classification models. In addition to speed, Agile Modeling aims to tackle the challenges posed by subjective vision models. As classification tasks become more nuanced, user interaction becomes increasingly crucial. However, it is important to note that the human-in-the-loop approach can be expensive due to the need of continuous human involvement and expertise. While this work aims at reducing time users spend on tuning their classification models, we propose an assisted method to automate parts of the pipeline and eliminate crowd-rater involvement.

\noindent\textbf{Vision-language models (VLMs).}
In the rapidly evolving domain of VLMs, two primary streams have emerged: contrastive and generative models. Contrastive models, such as CLIP \cite{radford2021learning} and ALIGN \cite{jia2021scaling}, leverage large-scale datasets to directly learn visual concepts from raw text, enabling high-accuracy zero-shot classification on open vocabularies \cite{he2016deep,deng2009imagenet}.
Generative models such as PaLI \cite{chen2022pali,chen2023pali,wang2023non,chen2023pali3} and GPT-V \cite{openai2023gpt4,openai2023gpt4v} focus on generating text from a combination of visual and text inputs. For instance, PaLI, trained on a vast collection of image-text pairs in various languages, achieves top performance across a range of vision and language tasks. Similarly, GPT-V allows the processing of image inputs, thereby enhancing the applicability of language models to multimodal tasks.
Other methods such as CoCa~\cite{yu2022coca,Toubal2023MultiModal} proposed a hybrid approach for simultaneously learning with generative and contrastive objectives. Despite their strength, VLMs capture visual data semantics, often prioritizing salient image features over nuanced visual cues. For instance, CLIP embeddings are intentionally compressed to encapsulate its most prominent subject \cite{shen2021much}.
Additionally, PaLI may struggle to provide detailed descriptions of complex scenes with numerous objects, as its training data predominantly lacks detailed annotations. In contrast, our proposed method is more stable and less sensitive to question phrasing as observed in \cref{fig:pali-vs-modeling-copilot}. 

\noindent\textbf{Large language models (LLMs) and tool-use.}
Large Language Models (LLMs) have revolutionized the landscape of artificial intelligence\cite{brown2020gpt3, devlin2018bert,touvron2023llama,falcon40b,refinedweb}, particularly in the field of natural language processing (NLP) and cognitive reasoning. By leveraging advanced methodologies such as chain-of-thought reasoning \cite{wei2022chain}, few-shot learning \cite{brown2020language,ouyang2022training}, and tool-use \cite{schick2023toolformer,hu2023reveal}, these models demonstrate exceptional performance across a wide spectrum of downstream tasks \cite{radford2019language}. They can operate across various modalities and a broad range of applications while maintaining high performance without the need for additional training.
Recent progress in integrating external tools with LLMs \cite{li2023lmeye,chen2023towards,you2023idealgpt,hu2023reveal,hu2023avis} has yielded systems like Toolformer \cite{schick2023toolformer}. This approach makes intelligent decisions about which APIs to invoke, optimizing the timing, arguments passed, and the subsequent assimilation of the results into future token predictions. This enhances zero-shot performance across a variety of tasks, establishing a solid foundation for LLMs to operate beyond their inherent capabilities. For fine-grained VQA, AVIS \cite{hu2023avis} introduces an autonomous information-seeking mechanism. By dynamically leveraging an LLM in tandem with external tools, it adeptly traverses a combinatorial search space. This is achieved through its unique approach of mimicking human decision-making processes, crafting a transition graph that guides the LLM's strategic decisions. Another tool-use enabled LLM system is ViperGPT \cite{suris2023vipergpt}, which embodies an innovative approach to tackling visual queries. It leverages a code-generation strategy that enables the seamless integration of vision-and-language models through the generation of Python code. This method, along with other similar methods (MMReact \cite{yang2023mm}, HuggingGPT \cite{shen2023hugginggpt}, Chameleon \cite{lu2023chameleon}, and Visual ChatGPT \cite{wu2023visual}) circumvents the need for extended training and ensures resilience across a diverse set of visual tasks. Collectively, these systems highlight the burgeoning synergy between LLMs and external tool use, pushing the frontiers of what LLMs can achieve. In our work, we adopt and extend ideas from these approaches to tackle the problem of subjective image classification.

\noindent\textbf{Customized prompts via language models.}
Customized Prompts via Language models (CuPL) \cite{pratt2022does} leverages CLIP's capabilities \cite{radford2021learning} to achieve zero-shot image classification. CuPL measures the similarity between an image and each visual class to perform classification. Typically, the classes are passed into CLIP's text encoder within a template such as ``photo of a bird'' for the class bird. CuPL employs GPT \cite{brown2020gpt3} to generate more comprehensive text descriptions for each class before feeding into CLIP.  This straightforward and zero-shot approach yields improved accuracy across various zero-shot image classification benchmarks. However, its evaluation has been limited to objective classification tasks and not on nuanced or subjective visual classification tasks. This approach for automatically annotating data improves upon CLIP but suffers from the same limitations compared to our work.

\section{Method} 
We propose an end-to-end system that streamlines the development of classifiers for nuanced visual concepts, addressing the limitations of traditional classifier development methods. The system consists of three core components, described in detail in the following subsections: (a) data mining, (b) annotation, (c) model training with active learning.

To build a classifier for a new concept, the user first provides a concept name and an optional description. The system then automatically mines images relevant to the concept and annotates them using a mixture of Large Language Models (LLM), Vision-Language Models (VLM), and Visual-Question-Answering (VQA) models. The annotated images are used to train a basis classification model, which is further refined through multiple rounds of active learning, resulting in a highly accurate classifier.

This setup mirrors the workflow of traditional classifier development, but it eliminates the need for costly and time-consuming human annotation which is a significant bottleneck in traditional methods. The Modeling Collaborator Annotator component, powered by LLMs and VLMs, enables zero-shot image labeling and drastically minimizes our dependence on user annotations.

\subsection{Data mining}
Mining quality data for training has traditionally been a labor-intensive process. This process begins with the clear definition of a concept, followed by the hunt for relevant images, and ends in the manual annotation of each of these images~\cite{deng2009imagenet,lin2014microsoft}. Particularly for nuanced visual tasks, there is a possibility that certain subtle visual patterns might be overlooked during data collection. Consequently, to ensure a comprehensive capture of all visual patterns, multiple iterations of refinement may be needed. In traditional Agile Modeling~\cite{stretcu2023agile} this challenge is addressed by soliciting \textit{users} to annotate data or generate new search queries to find more image examples. Each query results in a new semantic image search algorithm~\cite{jia2021scaling,radford2021learning} to gather other similar positive image examples for annotation from the public domain (LAION Dataset)~\cite{schuhmann2022laion}. 
Even with user intervention, \textit{user} queries may overlook essential cues, potentially leading to a deficit of hard negatives or a lack of coverage in specific visual modes. Additionally, the labels can vary between users, leading to potential human biases. 

To address human bias and minimize manual effort, we propose a data mining algorithm based on LLM chain-of-thought reasoning. While LLMs are not inherently unbiased\cite{gallegos2023bias} and may reflect biases present in their training data, they can assess a wider range of concepts at large scales from their extensive knowledge base, thus identifying a broader array of potential examples more efficiently.
First, we prompt the LLM to generate multiple positive and negative queries based on a concept's name and its description. Note that we do not directly assign images as positive or negative based on the query; rather, the goal is obtain representative images spanning both positive and hard-negative examples. To increase coverage and diversity, we expand the queries by instructing the LLM to apply various \textit{mutations}. For example, we may ask the LLM to iteratively come up with broader or narrower versions of the queries, or come up with variations for specific parts of the queries.
Drawing parallels to Agile Modeling, we use each query to extract image samples from the public domain~\cite{schuhmann2022laion}.

\label{sec:methods}

\begin{figure*}[ht]
    \centering
    \includegraphics[width=\linewidth]{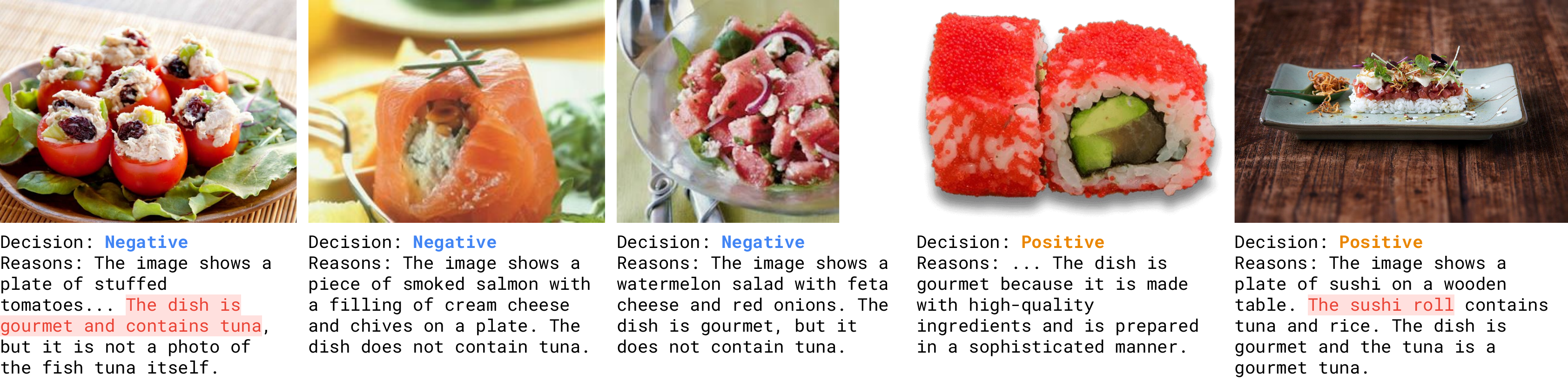}
    \includegraphics[width=\linewidth]{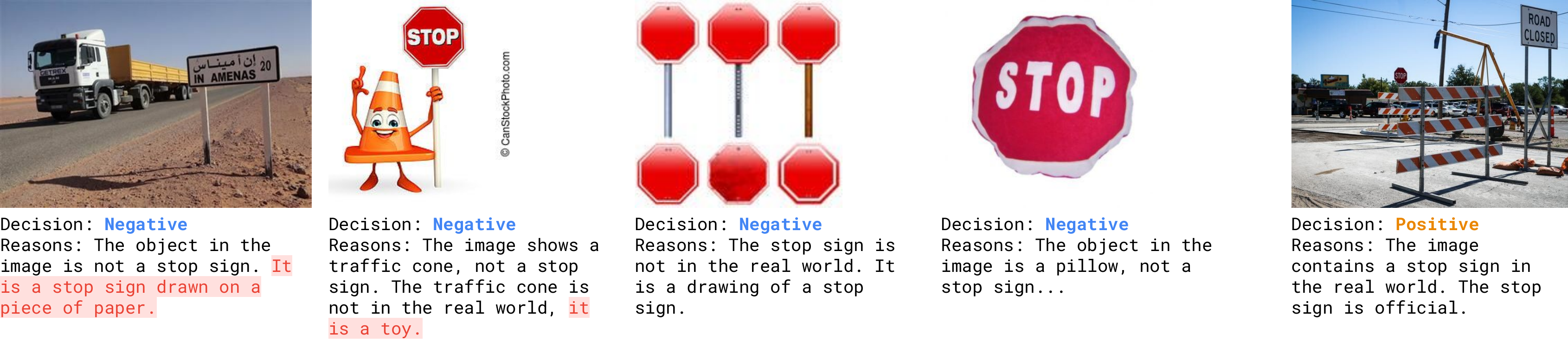}
    \caption{Modeling Collaborator Annotator examples for the concepts \texttt{gourmet tuna} (first row) and \texttt{stop sign} (second row). Hard negatives mined from the LAION dataset are shown in addition to some actual positives for the visual concept. The Modeling Collaborator Annotator is able to label the images as positive or negative as well as provide rationale. In some instances, the rationale could be incorrect (highlighted in red) due to error in VQA responses or hallucinations from the LLMs. Some of the reasons have been truncated for brevity.}
    \label{fig:results-modeling-copilot-examples}
\end{figure*}

\subsection{Modeling Collaborator Annotator}

\cref{fig:methods-annotation-process} describes the image annotation process. Our system effectively orchestrates the annotation process leveraging LLM's ability to invoke VLMs and other tools. It comprises three primary AI-driven modules: an LLM, a Captioning VLM \cite{wang2023non}, and a VQA VLM \cite{chen2023pali}. The automated annotation process is structured as follows:

\noindent \textbf{Concept initialization}: Initially, our system receives a concept name  (e.g., \texttt{gourmet tuna}), and optionally a concept description. If a concept description is absent, the LLM generates an initial description. This template can be modified by the user to cover all specifications and carve-outs.
    
\noindent \textbf{Attribute Extraction}: Based on the concept specifications, the LLM identifies objective attributes associated with the concept, such as ``image contains tuna'', ``is tuna sandwich'', and ``is tuna steak''.
    
\noindent \textbf{Attribute decomposition}: The LLM decomposes complex attributes into more granular and atomic attributes.
    
\noindent \textbf{Question generation}: The LLM then formulates a series of questions tailored for the VQA model. Examples include ``does the image contain food'', ``is the food tuna'', and ``is it tuna steak''.
    
\noindent \textbf{Visual assessment}: When an image is input, the VQA model processes these questions, yielding concise answers for each. Concurrently, the Captioning VLM provides a comprehensive description of the image.
    
\noindent \textbf{Final annotation}: With the textual data from the VLMs and the user's initial concept specification, the LLM employs chain-of-thought reasoning. It annotates the image as either positive or negative, also offering insights into its decision-making process.


Our approach utilizes the strengths of VLM, VQA, and LLM models while simultaneously avoiding their shortcomings. For example, VLMs, despite their capabilities, often struggle with nuanced and subjective concepts in classification tasks. Their performance hinges on the breadth and quality of training data, potentially leading to biases or gaps in understanding \cite{tan2019lxmert}. Ambiguities in language and the inherent subjectivity of certain questions can further challenge their accuracy \cite{lu2016hierarchical}. Moreover, these models, lacking real-world context and experiential understanding, might miss deeper cultural or emotional nuances \cite{goyal2017making}. Thus, while powerful, VLMs have inherent limitations in addressing intricate or subjective visual-linguistic tasks. \cref{fig:pali-vs-modeling-copilot} shows an example VLMs' (PaLI-X \cite{chen2023pali}) sensitivity to prompts.

VLMs are primarily designed for understanding and answering questions related to visual content, rather than performing deep chain-of-thought reasoning typical of advanced LLMs \cite{tan2019lxmert, lu2016hierarchical,wu2023autogen,press2022measuring}. While VLMs can comprehend simpler questions about images, they usually operate in a single-shot manner, providing answers based on the immediate visual and textual inputs without extended reasoning.
On the other hand, LLM question answering quality can be significantly improved through chain-of-thought reasoning, maintaining a coherent line of thought across extended text. Other techniques such as prompt chaining involve using a model's output as part of the subsequent input, simulating a sustained dialogue or iterative reasoning. Additionally, to extract deeper insights, users can guide LLMs with specific instructions, such as asking the model to think step-by-step \cite{yang2023large} or weigh pros and cons, thus simulating a more deliberate reasoning process \cite{brown2020gpt3}.

\subsection{Training and active learning}

While one could directly use the annotator as a model, this is prohibitive in many scenarios because of the high inference cost. For this reason, we adopt an approach similar to \cite{stretcu2023agile} for model training and active learning. Specifically, we first extract image features from a foundation vision model (CLIP or ALIGN) \cite{juan2019graph,jia2021scaling}. We then train a shallow multi-layer perceptron (MLP) with layer sizes $(128, 128, 128)$ to perform binary classification for the given concept. This can also be viewed as student-teacher distillation~\cite{hinton2015distilling} where we use the LLM-based annotator as the teacher model. We use a learning rate of $3\times10^{-4}$, a batch size of $512$, and optimize using AdamW \cite{loshchilov2018decoupled}.

After the initial model is trained, we perform multiple rounds of active learning. Each active-learning iteration consists of three stages. First, the lightweight classification model is applied to a large database of unlabeled images (LAION\cite{schuhmann2022laion}). Then, we perform stratified sampling to acquire candidate images for further AL rounds\cite{stretcu2023agile}. The intention is to capture hard negatives and hard positives that will boost precision and recall respectively. Second, our LLM-based annotator is autonomously applied to the selected images, providing additional training ground-truth. Thirdly, the student classifier is retrained, leveraging all the extant labeled data. We experiment with both margin sampling and stratified sampling techniques \cite{settles2009active} to mine examples during this active learning phase. The overall system thus adeptly balances between exploration (achieved via data mining through text search queries and expansion) and exploitation (achieved via active learning to mine visual modes that reduce model uncertainties).





\subsection{Implementation details}
As a large language model, we use PaLM 2 \cite{chowdhery2022palm,anil2023palm2} which was trained on a variety of different tasks, all of which helps PaLM 2 learn different aspects of language. Additionally, we use both the VQA and MMIT (multimodal instruction-tuned \cite{wang2023non}) variants of PaLI-X \cite{chen2023pali}. The particular choice of foundation models is based on their SOTA performance at the time of writing. These models have not been further trained or fine-tuned in this work.
\section{Experiments} 
\label{sec:results}

\begin{table*}
\centering
\resizebox{\linewidth}{!}{
\begin{tabular}{lrrrrrrrrrrrr}
\hline
& \multicolumn{3}{c}{\textbf{PaLI-X \cite{chen2023pali}}} & \multicolumn{3}{c}{\textbf{\hspace{0.3cm} CLIP \cite{radford2021learning}}} & \multicolumn{3}{c}{\textbf{\hspace{0.3cm} CuPL \cite{pratt2022does}}} & \multicolumn{3}{c}{\textbf{\hspace{0.3cm} Ours}} \\ \hline
\textbf{Concept}      & \textbf{Pre}  & \textbf{Rec}  & \textbf{F1} & \hspace{0.4cm} \textbf{Pre}    & \textbf{Rec}    & \textbf{F1}   & \hspace{0.4cm} \textbf{Pre}    & \textbf{Rec}    & \textbf{F1}   & \hspace{0.4cm} \textbf{Pre}   & \textbf{Rec}  & \textbf{F1}  \\ \hline \hline
\textbf{Easy concepts}             &      &      &      &      &      &      &      &      &      &      &      &      \\
\hspace{0.25cm} arts-and-crafts                    & 0.71 & 0.97 & 0.82 & 0.68 & 0.86 & 0.76 & 0.68 & 0.90 & 0.77 & 0.96 & 0.75 & 0.84 \\
\hspace{0.25cm} dance                              & 0.57 & 0.87 & 0.69 & 0.51 & 0.95 & 0.66 & 0.52 & 0.89 & 0.66 & 0.67 & 0.95 & 0.79 \\
\hspace{0.25cm} emergency-service                  & 0.67 & 0.88 & 0.76 & 0.53 & 0.87 & 0.65 & 0.54 & 0.91 & 0.67 & 0.88 & 0.73 & 0.76 \\
\hspace{0.25cm} hair-coloring                      & 0.76 & 0.97 & 0.85 & 0.70 & 0.99 & 0.82 & 0.70 & 0.99 & 0.82 & 0.76 & 0.97 & 0.85 \\
\hspace{0.25cm} in-ear-headphones                  & 0.70 & 0.96 & 0.81 & 0.43 & 0.95 & 0.59 & 0.44 & 0.96 & 0.60 & 0.82 & 0.86 & 0.82 \\
\hspace{0.25cm} pie-chart                          & 0.80 & 0.96 & 0.88 & 0.52 & 0.80 & 0.63 & 0.50 & 0.92 & 0.65 & 0.80 & 0.96 & 0.88 \\
\hspace{0.25cm} single-sneaker                     & 0.65 & 0.92 & 0.76 & 0.51 & 0.99 & 0.67 & 0.51 & 1.00 & 0.67 & 0.70 & 0.88 & 0.78 \\ \hline
\textbf{Easy concepts average}     & \underline{0.69} & \underline{0.93} & \underline{0.80} & 0.55 & 0.92 & 0.68 & 0.56 & \textbf{0.94} & 0.69 & \textbf{0.80} & 0.87 & \textbf{0.82} \\
$\Delta$                                   &      &      &      &      &      &      &      &      &      & +11\% & -6\% & +2\% \\ \hline
\textbf{Hard concepts}             &      &      &      &      &      &      &      &      &      &      &      &      \\
\hspace{0.25cm} astronaut                          & 0.61 & 0.87 & 0.71 & 0.40 & 0.95 & 0.56 & 0.42 & 0.95 & 0.58 & 0.72 & 0.79 & 0.72 \\
\hspace{0.25cm} block-tower                        & 0.45 & 0.97 & 0.62 & 0.38 & 0.99 & 0.55 & 0.37 & 0.98 & 0.54 & 0.89 & 0.68 & 0.66 \\
\hspace{0.25cm} gourmet-tuna                       & 0.52 & 0.95 & 0.67 & 0.29 & 1.00 & 0.45 & 0.29 & 1.00 & 0.45 & 0.52 & 0.95 & 0.67 \\
\hspace{0.25cm} hand-pointing                      & 0.56 & 0.99 & 0.71 & 0.39 & 0.87 & 0.54 & 0.39 & 0.94 & 0.55 & 0.89 & 0.79 & 0.74 \\
\hspace{0.25cm} healthy-dish                       & 0.38 & 1.00 & 0.55 & 0.37 & 0.99 & 0.54 & 0.38 & 1.00 & 0.55 & 0.84 & 0.61 & 0.61 \\
\hspace{0.25cm} home-fragrance                     & 0.57 & 0.51 & 0.54 & 0.40 & 0.95 & 0.56 & 0.40 & 0.96 & 0.57 & 0.57 & 0.51 & 0.54 \\
\hspace{0.25cm} stop-sign                          & 0.61 & 0.99 & 0.76 & 0.48 & 1.00 & 0.65 & 0.49 & 0.99 & 0.65 & 0.83 & 0.83 & 0.81 \\ \hline
\textbf{Hard concepts average}     & \underline{0.53} & \underline{0.90} & \underline{0.65} & 0.39 & 0.96 & 0.55 & 0.39 & \textbf{0.97} & 0.56 & \textbf{0.75} & 0.74 & \textbf{0.68} \\
$\Delta$                                   &      &      &      &      &      &      &      &      &      & +22\% & -16\% & +3\% \\ \hline
\textbf{Overall average}           & \underline{0.61} & \underline{0.92} & \underline{0.72} & 0.47 & 0.94 & 0.62 & 0.47 & \textbf{0.96} & 0.62 & \textbf{0.78} & 0.79 & \textbf{0.74} \\ 
$\Delta$                                   &      &      &      &      &      &      &      &      &      & +17\% & -13\% & +2\% \\ \hline \hline

\textbf{Hateful memes \cite{kiela2020hateful}}          & \underline{\textbf{0.66}} & \underline{0.42} & \underline{0.51} & 0.49 & \textbf{0.98} & 0.66 & 0.50 & 0.87 & 0.64 & 0.58 & 0.77 & \textbf{0.66} \\
$\Delta$                                   &      &      &      &      &      &      &      &      &      & -8\% & +35\% & +15\% \\ \hline 
\end{tabular}
}
\caption{Teacher performance (Precision, Recall, and F1 scores). Modeling Collaborator outperforms state-of-the-art zero-shot methods including CLIP, CuPL, and visual query answering models (PaLI-X). Underlined results represent the baseline (PaLI-X) with which our performance is compared to (deltas). We bold the best precision, recall, and F1 for easy concepts, hard concepts and Hateful memes dataset.}
\label{tab:results-teacher-performance}
\end{table*}

\begin{table*}[htpb]
\centering

\begin{tabular}{lrrrrrr}
\hline
                               & \multicolumn{3}{c}{Human Annotators}                    & \multicolumn{3}{c}{\hspace{1cm} Machine Annotators}               \\
\textbf{Concept}               & \textbf{User} & \textbf{Crowd} & \textbf{Crowd}        & \hspace{1cm} \textbf{CuPL} & \textbf{PaLI-X} & \textbf{Ours}         \\
Dataset size (per concept)     & $\sim$600       & $\sim$600      & $\sim$3000            & $\sim$3000    & $\sim$3000    & $\sim$3000            \\ \hline
\textbf{Easy concepts}         &                 &                &  &               &               &                       \\
arts-and-crafts                & 0.77            & 0.73           & 0.86                  & 0.78          & 0.77          & 0.78                  \\
dance                          & 0.69            & 0.70           & 0.81                  & 0.72          & 0.68          & 0.68                  \\
emergency-service              & 0.75            & 0.71           & 0.78                  & 0.59          & 0.66          & 0.72                  \\
hair-coloring                  & 0.85            & 0.85           & 0.83                  & 0.77          & 0.58          & 0.80                  \\
in-ear-headphones              & 0.73            & 0.66           & 0.67                  & 0.65          & 0.73          & 0.72                  \\
pie-chart                      & 0.77            & 0.76           & 0.76                  & 0.72          & 0.82          & 0.82                  \\
single-sneaker                         & 0.74            & 0.64           & 0.68                  & 0.51          & 0.61          & 0.56                  \\ \hline
\textbf{Easy concepts average} & 0.76            & \underline{0.72}     & 0.77                  & 0.68          & 0.69          & \textbf{0.73  (+1\%)} \\ \hline
\textbf{Hard concepts}         &                 &                &                       &               &               &                       \\
astronaut                      & 0.67            & 0.71           & 0.66                  & 0.60          & 0.65          & 0.65                  \\
block-tower                    & 0.59            & 0.58           & 0.45                  & 0.48          & 0.49          & 0.50                  \\
gourmet-tuna                   & 0.50            & 0.51           & 0.35                  & 0.54          & 0.52          & 0.52                  \\
hand-pointing                  & 0.50            & 0.56           & 0.58                  & 0.56          & 0.81          & 0.81                  \\
healthy-dish                   & 0.59            & 0.49           & 0.47                  & 0.42          & 0.45          & 0.53                  \\
home-fragrance                 & 0.62            & 0.60           & 0.69                  & 0.56          & 0.53          & 0.53                  \\
stop-sign                      & 0.70            & 0.57           & 0.55                  & 0.62          & 0.51          & 0.64                  \\ \hline
\textbf{Hard concepts average} & 0.60            & \underline{0.57}     & 0.54                  & 0.54          & 0.57          & \textbf{0.60  (+3\%)} \\ \hline
\textbf{Overall average}       & 0.68            & \underline{0.65}     & 0.65                  & 0.61          & 0.63          & \textbf{0.66  (+1\%)} \\ \hline
\end{tabular}

\caption{Quality comparison of different annotators (or teacher models) using the final distilled model performance (auPR). Concept owners provide the highest quality annotations because of their deep understanding of the nuanced concept. Modeling Collaborator annotator provides better quality labels compared with labor-intensive annotations from crowd raters, and compared to other automated methods.}
\label{tab:human-vs-machine-annotators}
\end{table*}

\begin{figure*}[ht]
    \centering
    \includegraphics[width=\linewidth]{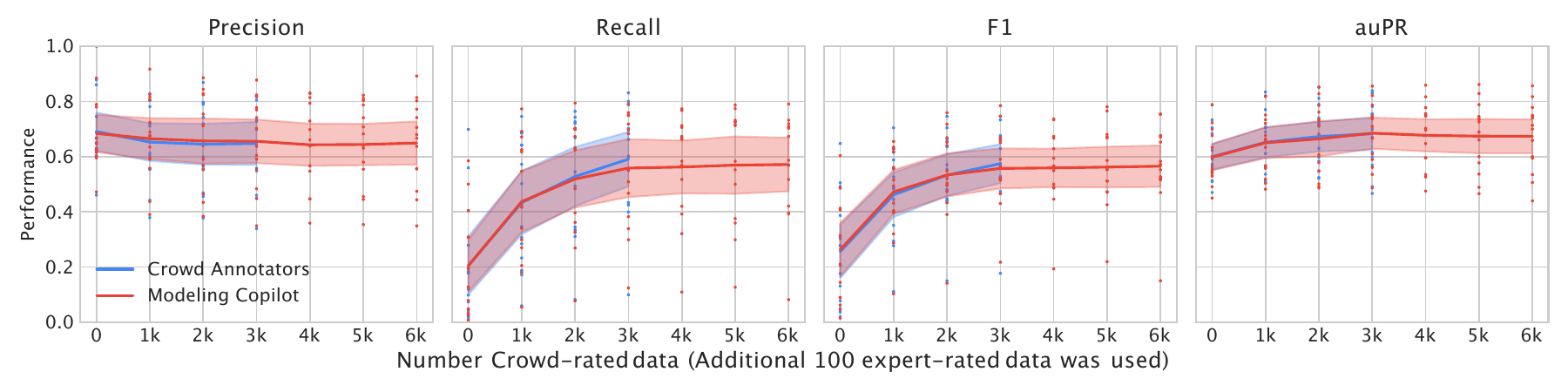}
    \vspace{-0.7cm}
    \caption{Comparing the contribution of increasingly more training examples annotated by crowd-annotators vs. Modeling Collaborator Annotator (fully automated). The y-axis shows the performance of the final distilled model. When \emph{user} feedback is minimal (100 annotated examples), more crowd-annotators examples improve the final distilled model despite the noisy prediction. Modeling Collaborator Annotator provides similar improvement of performance without any human interactions and can be scaled better to annotate a lot more examples due to its autonomy.}
    \label{fig:results-crowd-vs-copilot}
\end{figure*}


We present our experimental setup and results with three takeaways. First, we show that \ours Annotator outperforms other zero-shot methods (CLIP \cite{radford2021learning}, CuPL\cite{pratt2022does} and PaLI-X\cite{chen2023pali}). Second, while \ours Annotator is able to beat state-of-the-art methods in both easy and hard concepts, we see much larger gains on harder and more subjective concepts. Finally, when using our end to end system, we can produce deployable models of competitive quality with minimal \emph{user} annotations (100 annotations vs. 2,000 in traditional Agile Modeling). 

\noindent \textbf{Datasets.} In addition to the LAION dataset used for data mining in our system, we evaluate our methods on the public Hateful Memes dataset~\cite{kiela2020hateful}. For evaluation and user-study, we use the Agile Modeling dataset \cite{stretcu2023agile} that is comprised of 14 concepts, each with positive and negative images mined from the LAION dataset. This dataset is split into \emph{easy} and \emph{hard} concepts depending on the zero-shot performance on each concept using CLIP as described in \cite{stretcu2023agile}.

\noindent \textbf{Models.} We benchmark \ours Annotator against state-of-the-art zero-shot and open-vocabulary classifiers: CLIP \cite{radford2021learning}, CuPL\cite{pratt2022does}, and PaLI-X (55B) \cite{chen2023pali} as a generative VQA model. We evaluate CLIP by embedding the name of the concept and measuring the cosine similarity to each image embedding. CuPL uses the same technique but instead of embedding the concept name directly, we embed a description of the concept generated by an LLM. Both GPT3 and PaLM 2 models were experimented with but we chose PaLM 2 since it produced superior results. In the case of CLIP and CuPL, we select an operating point using a grid search maximizing the F1 score on a subset of the training set. We use PaLI-X VQA variant as a classifier by prompting it \emph{``Is this an image of $\mathcal{X}$?''} and we assign a positive or negative prediction based on its answer.

\noindent \textbf{Annotator Adaptation.} While testing the system, we observed some amount of concept-dependent variability in the Annotator. For example, for simple concepts like ``cat'' a VLM might already have state-of-the-art performance and our system can even degrade quality in these cases. To address this we implemented six different Annotator \emph{strategies}. While developing a classifier for a particular concept, we have the concept owner build an on-the-fly validation set of 100 images which is then used to select the best performing strategy for that particular concept. Different parameters describing these configurations are explained in the \supplementary.

\noindent \textbf{Users, Crowd, and \ours.} We measure the agreement/alignment with the \emph{user} for both the crowd and automatic annotation methods. The \emph{user} is the source of ground-truth and the person manually annotating the test set. \emph{Crowd} annotators are given a description and examples by the \emph{user} and asked to annotate images at a larger scale. \emph{\ours} Annotator is able to scale up the annotation process further due to its autonomy and can encapsulate an image set of higher diversity in visual modes. We measure the annotator alignment by comparing the performance (auPR) on the distilled model trained on data annotated by different human and machine annotators. 


\subsection{Modeling Collaborator Annotator}
\noindent \textbf{\ours Annotator outperforms other zero-shot methods}. We show the results of these experiments in \cref{tab:results-teacher-performance}. We measure the alignment with the \emph{user} on the held-out test set of the Agile Modeling dataset using agreement scores (precision, recall, and F1). CLIP and CuPL contrastive models suffer from very low precision in favor of high recall. PaLI-X outperforms contrastive models, making it more suitable as a baseline for our proposed Annotator.
\textbf{We achieve significant gains for subjective (hard) concepts while maintaining equivalent performance for less subjective (easy) concepts.}
\cref{tab:results-teacher-performance} shows a significant skew in concept improvement: over 25\% of the concepts showed an F1 score gain of 4\% or higher, including \texttt{hateful memes} \cite{kiela2020hateful} at 15\%, \texttt{healthy-dish} at 6\%, and \texttt{stop-sign} at 5\%, exhibiting substantial improvements in areas requiring more subjective classifications. 
This trend indicates that our model is particularly effective for complex or subjective concepts, but may offer only marginal benefits for concepts that PaLI-X is already good at.
Regardless, a Wilcoxon Signed-Rank Test on the F1 scores comparing our system against PaLI-X yields a statistically significant improvement across all concepts ($p < 0.01$). In addition to classification, our system outputs rationales shown in \cref{fig:results-modeling-copilot-examples}.

\begin{table}[ht]
\centering
\resizebox{\linewidth}{!}{
\begin{tabular}{lcrrrrr}
\hline
\textbf{Method}                 & \textbf{Labeler} & \# \textbf{Ex}. & \textbf{F1}   & \textbf{Acc}  & \textbf{Pre}  & \textbf{Rec}  \\ \hline
Ours (Teacher)                  & -         & - & 0.66 & 0.61 & 0.58 & 0.77 \\ \hline \hline
CLIP \cite{radford2021learning} & - & - & \textbf{0.57} & 0.53 & 0.51 & 0.65 \\
CuPL \cite{pratt2022does}       & - & - & 0.51 & \textbf{0.64} & 0.50 & \textbf{0.87}      \\
PaLI-X \cite{chen2023pali}      & - & - & 0.51 & 0.61 & \textbf{0.66} & 0.42 \\
Ours (Student)                  & MC & 7K & 0.56 & 0.52 & 0.50 & 0.64     \\ \hline
CLIP+MLP                        & Human     & 8.5K & 0.48 & 0.60 & 0.65 & 0.38 \\ \hline
\end{tabular}}
\caption{Performance of our method (both Annotator and distilled models) on the Hateful Memes \cite{kiela2020hateful} public dataset. Zero-shot and VQA methods are used for comparison.}
\label{tab:results-hatefulmemes}
\end{table}

\subsection{Human-machine alignment}

\noindent \textbf{\ours can produce deployable models of competitive quality with minimal user annotations.}
We measure the effect of using varying levels of human and automated annotation in \cref{tab:human-vs-machine-annotators}.
We note that, while our model cannot exceed the distilled user model performance (distilled on 100\% accurate annotations), we can outperform crowd-raters. Our Annotator system significantly outperforms crowd-raters on harder more nuanced concepts (different of 6\%). Whereas it slightly under-performs on easy concepts. This is likely due to prediction errors from automated VQA models (PaLI-X) where humans show better performance.
In comparison to using other state-of-the-art open-vocabulary zero-shot annotators (CLIP, CuPL and PaLI-X), our system outperforms these methods on both easy and hard concepts.
Our fully automated system successfully generates distilled models that match the quality of ones crafted with classical Agile Modeling, with performance within a 2\% margin of the \emph{user}'s output. 
\cref{fig:results-crowd-vs-copilot} shows that both crowd-annotators and \ours Annotator can improve the performance of the distilled model, even when \emph{user} feedback is minimal. However, \ours Annotator has the advantage of being fully automated and can scale to a larger number of examples.

\noindent \textbf{\ours and other zero-shot and classical methods fail in complex visual tasks that require complex understanding and reasoning.}
The effectiveness of our method on identifying hateful memes \cite{kiela2020hateful}, as demonstrated in  \cref{tab:results-hatefulmemes}, is further highlighted by its ability to match fully-trained models without relying on labeled data. Both the teacher and student models outperform the traditional training approach without using any of the training datasets. However, the performance is still low, demonstrating the limitations of our approach.

\section{Limitations} 
\label{sec:limitations}

As our system is an orchestration of LLMs and VLMs, it can suffer from some of the limitations of its atomic components (PaLM 2, PaLI-X, and CLIP). For example, we observed that providing verbose and overly-complex descriptions of simple concepts (cats, dogs, etc.) can actually degrade performance in comparison to simply using PaLI-X.
Another issue is that for certain concepts, the CLIP features can lead to poor distilled model quality. One example is \texttt{stop sign} (where the stop sign is expected to be a real stop sign in traffic), where the CLIP feature could capture the overall semantics of stop signs, but could not easily discriminate between physical instances vs depictions.
\section{Conclusion} 
\label{sec:conclusion}
In this paper, we presented Modeling Collaborator, a novel framework that alleviates the manual effort required to develop classifiers for subjective and nuanced visual concepts. Our framework leverages advancements in large language models (LLMs) and vision-language models (VLMs) to carve out the concept space through conversation and by automatically labeling training data points. We demonstrate the effectiveness of our framework through a set of experiments, showing that it can quickly build visual classifiers for nuanced concepts and outperform both traditional Agile Modeling and state-of-the-art zero-shot classification models. Our work has the potential to significantly reduce the time and effort required to develop classifiers for a wide range of applications including content moderation and aesthetic classification.

{\small 
\bibliographystyle{ieee_fullname}
\bibliography{egbib}

\begin{thebibliography}{10}\itemsep=-1pt

\bibitem{falcon40b}
Ebtesam Almazrouei, Hamza Alobeidli, Abdulaziz Alshamsi, Alessandro Cappelli,
  Ruxandra Cojocaru, Merouane Debbah, Etienne Goffinet, Daniel Heslow, Julien
  Launay, Quentin Malartic, Badreddine Noune, Baptiste Pannier, and Guilherme
  Penedo.
\newblock {Falcon-40B}: an open large language model with state-of-the-art
  performance.
\newblock 2023.

\bibitem{anil2023palm2}
Rohan Anil, Andrew~M Dai, Orhan Firat, Melvin Johnson, Dmitry Lepikhin,
  Alexandre Passos, Siamak Shakeri, Emanuel Taropa, Paige Bailey, Zhifeng Chen,
  et~al.
\newblock Palm 2 technical report.
\newblock {\em arXiv preprint arXiv:2305.10403}, 2023.

\bibitem{brown2020gpt3}
Tom Brown, Benjamin Mann, Nick Ryder, Melanie Subbiah, Jared~D Kaplan, Prafulla
  Dhariwal, Arvind Neelakantan, Pranav Shyam, Girish Sastry, Amanda Askell,
  et~al.
\newblock Language models are few-shot learners.
\newblock {\em NeurIPS}, 33:1877--1901, 2020.

\bibitem{brown2020language}
Tom Brown, Benjamin Mann, Nick Ryder, Melanie Subbiah, Jared~D Kaplan, Prafulla
  Dhariwal, Arvind Neelakantan, Pranav Shyam, Girish Sastry, Amanda Askell,
  et~al.
\newblock Language models are few-shot learners.
\newblock {\em NeurIPS}, 33:1877--1901, 2020.

\bibitem{chen2023towards}
Liang Chen, Yichi Zhang, Shuhuai Ren, Haozhe Zhao, Zefan Cai, Yuchi Wang,
  Tianyu Liu, and Baobao Chang.
\newblock Towards end-to-end embodied decision making with multi-modal large
  language model: Explorations with gpt4-vision and beyond.
\newblock In {\em NeurIPS 2023 Foundation Models for Decision Making Workshop},
  2023.

\bibitem{chen2023pali}
Xi Chen, Josip Djolonga, Piotr Padlewski, Basil Mustafa, Soravit Changpinyo,
  Jialin Wu, Carlos~Riquelme Ruiz, Sebastian Goodman, Xiao Wang, Yi Tay, et~al.
\newblock Pali-x: On scaling up a multilingual vision and language model.
\newblock {\em arXiv preprint arXiv:2305.18565}, 2023.

\bibitem{chen2023pali3}
Xi Chen, Xiao Wang, Lucas Beyer, Alexander Kolesnikov, Jialin Wu, Paul
  Voigtlaender, Basil Mustafa, Sebastian Goodman, Ibrahim Alabdulmohsin, Piotr
  Padlewski, et~al.
\newblock Pali-3 vision language models: Smaller, faster, stronger.
\newblock {\em arXiv preprint arXiv:2310.09199}, 2023.

\bibitem{chen2022pali}
Xi Chen, Xiao Wang, Soravit Changpinyo, AJ Piergiovanni, Piotr Padlewski,
  Daniel Salz, Sebastian Goodman, Adam Grycner, Basil Mustafa, Lucas Beyer,
  et~al.
\newblock Pali: A jointly-scaled multilingual language-image model.
\newblock In {\em ICLR}, 2022.

\bibitem{chowdhery2022palm}
Aakanksha Chowdhery, Sharan Narang, Jacob Devlin, Maarten Bosma, Gaurav Mishra,
  Adam Roberts, Paul Barham, Hyung~Won Chung, Charles Sutton, Sebastian
  Gehrmann, et~al.
\newblock Palm: Scaling language modeling with pathways.
\newblock {\em Journal of Machine Learning Research}, 24(240):1--113, 2023.

\bibitem{connie2017smart}
Tee Connie, Mundher Al-Shabi, and Michael Goh.
\newblock Smart content recognition from images using a mixture of
  convolutional neural networks.
\newblock In {\em IT Convergence and Security 2017: Volume 1}, pages 11--18.
  Springer, 2017.

\bibitem{deng2009imagenet}
Jia Deng, Wei Dong, Richard Socher, Li-Jia Li, Kai Li, and Li Fei-Fei.
\newblock Imagenet: A large-scale hierarchical image database.
\newblock In {\em CVPR}, pages 248--255, 2009.

\bibitem{devlin2018bert}
Jacob Devlin, Ming-Wei Chang, Kenton Lee, and Kristina Toutanova.
\newblock Bert: Pre-training of deep bidirectional transformers for language
  understanding.
\newblock In {\em NAACL-HLT}, pages 4171--4186, 2019.

\bibitem{fiske1991social}
Susan~T Fiske and Shelley~E Taylor.
\newblock {\em Social cognition}.
\newblock Mcgraw-Hill Book Company, 1991.

\bibitem{frege1879begriffsschrift}
Gottlob Frege et~al.
\newblock Begriffsschrift, a formula language, modeled upon that of arithmetic,
  for pure thought.
\newblock {\em From Frege to G{\"o}del: A source book in mathematical logic},
  1931:1--82, 1879.

\bibitem{gallegos2023bias}
Isabel~O Gallegos, Ryan~A Rossi, Joe Barrow, Md~Mehrab Tanjim, Sungchul Kim,
  Franck Dernoncourt, Tong Yu, Ruiyi Zhang, and Nesreen~K Ahmed.
\newblock Bias and fairness in large language models: A survey.
\newblock {\em arXiv preprint arXiv:2309.00770}, 2023.

\bibitem{goyal2017making}
Yash Goyal, Tejas Khot, Douglas Summers-Stay, Dhruv Batra, and Devi Parikh.
\newblock Making the v in vqa matter: Elevating the role of image understanding
  in visual question answering.
\newblock In {\em CVPR}, pages 6904--6913, 2017.

\bibitem{he2016deep}
Kaiming He, Xiangyu Zhang, Shaoqing Ren, and Jian Sun.
\newblock Deep residual learning for image recognition.
\newblock In {\em CVPR}, pages 770--778, 2016.

\bibitem{hinton2015distilling}
Geoffrey Hinton, Oriol Vinyals, and Jeff Dean.
\newblock Distilling the knowledge in a neural network.
\newblock {\em arXiv preprint arXiv:1503.02531}, 2015.

\bibitem{hsieh2023tool}
Cheng-Yu Hsieh, Si-An Chen, Chun-Liang Li, Yasuhisa Fujii, Alexander Ratner,
  Chen-Yu Lee, Ranjay Krishna, and Tomas Pfister.
\newblock Tool documentation enables zero-shot tool-usage with large language
  models.
\newblock {\em arXiv preprint arXiv:2308.00675}, 2023.

\bibitem{hu2023avis}
Ziniu Hu, Ahmet Iscen, Chen Sun, Kai-Wei Chang, Yizhou Sun, David~A Ross,
  Cordelia Schmid, and Alireza Fathi.
\newblock Avis: Autonomous visual information seeking with large language
  models.
\newblock {\em arXiv preprint arXiv:2306.08129}, 2023.

\bibitem{hu2023reveal}
Ziniu Hu, Ahmet Iscen, Chen Sun, Zirui Wang, Kai-Wei Chang, Yizhou Sun,
  Cordelia Schmid, David~A Ross, and Alireza Fathi.
\newblock Reveal: Retrieval-augmented visual-language pre-training with
  multi-source multimodal knowledge memory.
\newblock In {\em CVPR}, pages 23369--23379, 2023.

\bibitem{ji2020action}
Jingwei Ji, Ranjay Krishna, Li Fei-Fei, and Juan~Carlos Niebles.
\newblock Action genome: Actions as compositions of spatio-temporal scene
  graphs.
\newblock In {\em CVPR}, pages 10236--10247, 2020.

\bibitem{jia2021scaling}
Chao Jia, Yinfei Yang, Ye Xia, Yi-Ting Chen, Zarana Parekh, Hieu Pham, Quoc Le,
  Yun-Hsuan Sung, Zhen Li, and Tom Duerig.
\newblock Scaling up visual and vision-language representation learning with
  noisy text supervision.
\newblock In {\em ICML}, pages 4904--4916, 2021.

\bibitem{juan2019graph}
Da-Cheng Juan, Chun-Ta Lu, Zhen Li, Futang Peng, Aleksei Timofeev, Yi-Ting
  Chen, Yaxi Gao, Tom Duerig, Andrew Tomkins, and Sujith Ravi.
\newblock Graph-rise: Graph-regularized image semantic embedding.
\newblock {\em arXiv preprint arXiv:1902.10814}, 2019.

\bibitem{khosla2015understanding}
Aditya Khosla, Akhil~S Raju, Antonio Torralba, and Aude Oliva.
\newblock Understanding and predicting image memorability at a large scale.
\newblock {\em ICCV}, pages 2390--2398, 2015.

\bibitem{kiela2020hateful}
Douwe Kiela, Hamed Firooz, Aravind Mohan, Vedanuj Goswami, Amanpreet Singh,
  Pratik Ringshia, and Davide Testuggine.
\newblock The hateful memes challenge: Detecting hate speech in multimodal
  memes.
\newblock {\em NeurIPS}, 33:2611--2624, 2020.

\bibitem{krishna2017visual}
Ranjay Krishna, Yuke Zhu, Oliver Groth, Justin Johnson, Kenji Hata, Joshua
  Kravitz, Stephanie Chen, Yannis Kalantidis, Li-Jia Li, David~A Shamma, et~al.
\newblock Visual genome: Connecting language and vision using crowdsourced
  dense image annotations.
\newblock {\em IJCV}, 123(1):32--73, 2017.

\bibitem{kuznetsova2020open}
Alina Kuznetsova, Hassan Rom, Neil Alldrin, Jasper Uijlings, Ivan Krasin, Jordi
  Pont-Tuset, Shahab Kamali, Stefan Popov, Matteo Malloci, Alexander
  Kolesnikov, et~al.
\newblock The open images dataset v4.
\newblock {\em ICCV}, 128:1956--1981, 2020.

\bibitem{li2023lmeye}
Yunxin Li, Baotian Hu, Xinyu Chen, Lin Ma, and Min Zhang.
\newblock Lmeye: An interactive perception network for large language models.
\newblock {\em arXiv preprint arXiv:2305.03701}, 2023.

\bibitem{lin2014microsoft}
Tsung-Yi Lin, Michael Maire, Serge Belongie, James Hays, Pietro Perona, Deva
  Ramanan, Piotr Doll{\'a}r, and C~Lawrence Zitnick.
\newblock Microsoft coco: Common objects in context.
\newblock In {\em ECCV}, pages 740--755, 2014.

\bibitem{loshchilov2018decoupled}
Ilya Loshchilov and Frank Hutter.
\newblock Decoupled weight decay regularization.
\newblock In {\em ICLR}, 2018.

\bibitem{lu2016visual}
Cewu Lu, Ranjay Krishna, Michael Bernstein, and Li Fei-Fei.
\newblock Visual relationship detection with language priors.
\newblock In {\em ECCV}, pages 852--869, 2016.

\bibitem{lu2016hierarchical}
Jiasen Lu, Jianwei Yang, Dhruv Batra, and Devi Parikh.
\newblock Hierarchical question-image co-attention for visual question
  answering.
\newblock {\em NeurIPS}, 29, 2016.

\bibitem{lu2023chameleon}
Pan Lu, Baolin Peng, Hao Cheng, Michel Galley, Kai-Wei Chang, Ying~Nian Wu,
  Song-Chun Zhu, and Jianfeng Gao.
\newblock Chameleon: Plug-and-play compositional reasoning with large language
  models.
\newblock {\em NeurIPS}, 36, 2024.

\bibitem{ma2023crepe}
Zixian Ma, Jerry Hong, Mustafa~Omer Gul, Mona Gandhi, Irena Gao, and Ranjay
  Krishna.
\newblock Crepe: Can vision-language foundation models reason compositionally?
\newblock In {\em CVPR}, pages 10910--10921, 2023.

\bibitem{miller1956magical}
George~A Miller.
\newblock The magical number seven, plus or minus two: Some limits on our
  capacity for processing information.
\newblock {\em Psychological review}, 63(2):81, 1956.

\bibitem{openai2023gpt4}
OpenAI.
\newblock Gpt-4 technical report, 2023.

\bibitem{openai2023gpt4v}
OpenAI.
\newblock Gpt-4v(ision) system card.
\newblock \url{https://cdn.openai.com/papers/GPTV_System_Card.pdf}, 2023.
\newblock Accessed: 2023-11-15.

\bibitem{ouyang2022training}
Long Ouyang, Jeffrey Wu, Xu Jiang, Diogo Almeida, Carroll Wainwright, Pamela
  Mishkin, Chong Zhang, Sandhini Agarwal, Katarina Slama, Alex Ray, et~al.
\newblock Training language models to follow instructions with human feedback.
\newblock {\em NeurIPS}, 35:27730--27744, 2022.

\bibitem{refinedweb}
Guilherme Penedo, Quentin Malartic, Daniel Hesslow, Ruxandra Cojocaru,
  Alessandro Cappelli, Hamza Alobeidli, Baptiste Pannier, Ebtesam Almazrouei,
  and Julien Launay.
\newblock The {R}efined{W}eb dataset for {F}alcon {LLM}: outperforming curated
  corpora with web data, and web data only.
\newblock {\em arXiv preprint arXiv:2306.01116}, 2023.

\bibitem{pratt2022does}
Sarah Pratt, Ian Covert, Rosanne Liu, and Ali Farhadi.
\newblock What does a platypus look like? generating customized prompts for
  zero-shot image classification.
\newblock In {\em Proceedings of the IEEE/CVF International Conference on
  Computer Vision}, pages 15691--15701, 2023.

\bibitem{press2022measuring}
Ofir Press, Muru Zhang, Sewon Min, Ludwig Schmidt, Noah~A Smith, and Mike
  Lewis.
\newblock Measuring and narrowing the compositionality gap in language models.
\newblock {\em arXiv preprint arXiv:2210.03350}, 2022.

\bibitem{radford2021learning}
Alec Radford, Jong~Wook Kim, Chris Hallacy, Aditya Ramesh, Gabriel Goh,
  Sandhini Agarwal, Girish Sastry, Amanda Askell, Pamela Mishkin, Jack Clark,
  et~al.
\newblock Learning transferable visual models from natural language
  supervision.
\newblock In {\em ICML}, pages 8748--8763, 2021.

\bibitem{radford2019language}
Alec Radford, Jeffrey Wu, Rewon Child, David Luan, Dario Amodei, Ilya
  Sutskever, et~al.
\newblock Language models are unsupervised multitask learners.
\newblock {\em OpenAI blog}, 1:9, 2019.

\bibitem{roy2017automated}
Arpita Roy, Anamika Paul, Hamed Pirsiavash, and Shimei Pan.
\newblock Automated detection of substance use-related social media posts based
  on image and text analysis.
\newblock In {\em 2017 IEEE 29th International Conference on Tools with
  Artificial Intelligence (ICTAI)}, pages 772--779. IEEE, 2017.

\bibitem{schick2023toolformer}
Timo Schick, Jane Dwivedi-Yu, Roberto Dess{\`\i}, Roberta Raileanu, Maria
  Lomeli, Luke Zettlemoyer, Nicola Cancedda, and Thomas Scialom.
\newblock Toolformer: Language models can teach themselves to use tools.
\newblock {\em arXiv preprint arXiv:2302.04761}, 2023.

\bibitem{schuhmann2022laion}
Christoph Schuhmann, Romain Beaumont, Richard Vencu, Cade Gordon, Ross
  Wightman, Mehdi Cherti, Theo Coombes, Aarush Katta, Clayton Mullis, Mitchell
  Wortsman, et~al.
\newblock Laion-5b: An open large-scale dataset for training next generation
  image-text models.
\newblock {\em NeurIPS}, 35:25278--25294, 2022.

\bibitem{settles2009active}
Burr Settles.
\newblock Active learning literature survey.
\newblock 2009.

\bibitem{shen2021much}
Sheng Shen, Liunian~Harold Li, Hao Tan, Mohit Bansal, Anna Rohrbach, Kai-Wei
  Chang, Zhewei Yao, and Kurt Keutzer.
\newblock How much can clip benefit vision-and-language tasks?
\newblock In {\em ICLR}, 2021.

\bibitem{shen2023hugginggpt}
Yongliang Shen, Kaitao Song, Xu Tan, Dongsheng Li, Weiming Lu, and Yueting
  Zhuang.
\newblock Hugginggpt: Solving ai tasks with chatgpt and its friends in hugging
  face.
\newblock {\em NeurIPS}, 36, 2024.

\bibitem{stretcu2023agile}
Otilia Stretcu, Edward Vendrow, Kenji Hata, Krishnamurthy Viswanathan, Vittorio
  Ferrari, Sasan Tavakkol, Wenlei Zhou, Aditya Avinash, Enming Luo, Neil~Gordon
  Alldrin, et~al.
\newblock Agile modeling: Image classification with domain experts in the loop.
\newblock {\em ICCV}, 2023.

\bibitem{suris2023vipergpt}
D{\'\i}dac Sur{\'\i}s, Sachit Menon, and Carl Vondrick.
\newblock Vipergpt: Visual inference via python execution for reasoning.
\newblock {\em arXiv preprint arXiv:2303.08128}, 2023.

\bibitem{tan2019lxmert}
Hao Tan and Mohit Bansal.
\newblock Lxmert: Learning cross-modality encoder representations from
  transformers.
\newblock In {\em Proceedings of the 2019 Conference on Empirical Methods in
  Natural Language Processing and the 9th International Joint Conference on
  Natural Language Processing (EMNLP-IJCNLP)}. Association for Computational
  Linguistics, 2019.

\bibitem{Toubal2023MultiModal}
Imad~Eddine Toubal, Yi-Ting Chen, Krishnamurthy Viswanathan, Daniel Salz, Ye
  Xia, and Zhongli Ding.
\newblock Multi-modal dual-tower architectures for entity retrieval from image
  and text.
\newblock In {\em CVPRW}, 2023.

\bibitem{touvron2023llama}
Hugo Touvron, Louis Martin, Kevin Stone, Peter Albert, Amjad Almahairi, Yasmine
  Babaei, Nikolay Bashlykov, Soumya Batra, Prajjwal Bhargava, Shruti Bhosale,
  et~al.
\newblock Llama 2: Open foundation and fine-tuned chat models.
\newblock {\em arXiv preprint arXiv:2307.09288}, 2023.

\bibitem{wang2023non}
Yaqing Wang, Jialin Wu, Tanmaya Dabral, Jiageng Zhang, Geoff Brown, Chun-Ta Lu,
  Frederick Liu, Yi Liang, Bo Pang, Michael Bendersky, et~al.
\newblock Non-intrusive adaptation: Input-centric parameter-efficient
  fine-tuning for versatile multimodal modeling.
\newblock {\em arXiv preprint arXiv:2310.12100}, 2023.

\bibitem{wei2022chain}
Jason Wei, Xuezhi Wang, Dale Schuurmans, Maarten Bosma, Fei Xia, Ed Chi, Quoc~V
  Le, Denny Zhou, et~al.
\newblock Chain-of-thought prompting elicits reasoning in large language
  models.
\newblock {\em NeurIPS}, 35:24824--24837, 2022.

\bibitem{wu2023visual}
Chenfei Wu, Shengming Yin, Weizhen Qi, Xiaodong Wang, Zecheng Tang, and Nan
  Duan.
\newblock Visual chatgpt: Talking, drawing and editing with visual foundation
  models.
\newblock {\em arXiv preprint arXiv:2303.04671}, 2023.

\bibitem{wu2023autogen}
Qingyun Wu, Gagan Bansal, Jieyu Zhang, Yiran Wu, Shaokun Zhang, Erkang Zhu,
  Beibin Li, Li Jiang, Xiaoyun Zhang, and Chi Wang.
\newblock Autogen: Enabling next-gen llm applications via multi-agent
  conversation framework.
\newblock {\em arXiv preprint arXiv:2308.08155}, 2023.

\bibitem{yang2023large}
Chengrun Yang, Xuezhi Wang, Yifeng Lu, Hanxiao Liu, Quoc~V Le, Denny Zhou, and
  Xinyun Chen.
\newblock Large language models as optimizers.
\newblock {\em arXiv preprint arXiv:2309.03409}, 2023.

\bibitem{yang2023mm}
Zhengyuan Yang, Linjie Li, Jianfeng Wang, Kevin Lin, Ehsan Azarnasab, Faisal
  Ahmed, Zicheng Liu, Ce Liu, Michael Zeng, and Lijuan Wang.
\newblock Mm-react: Prompting chatgpt for multimodal reasoning and action.
\newblock {\em arXiv preprint arXiv:2303.11381}, 2023.

\bibitem{you2023idealgpt}
Haoxuan You, Rui Sun, Zhecan Wang, Long Chen, Gengyu Wang, Hammad~A Ayyubi,
  Kai-Wei Chang, and Shih-Fu Chang.
\newblock Idealgpt: Iteratively decomposing vision and language reasoning via
  large language models.
\newblock {\em arXiv preprint arXiv:2305.14985}, 2023.

\bibitem{yu2022coca}
Jiahui Yu, Zirui Wang, Vijay Vasudevan, Legg Yeung, Mojtaba Seyedhosseini, and
  Yonghui Wu.
\newblock Coca: Contrastive captioners are image-text foundation models.
\newblock {\em arXiv preprint arXiv:2205.01917}, 2022.

\end{thebibliography}
}

\section*{Appendix}
\begin{appendix}
\section{Concept names and descriptions}
\subsection{Agile Modeling dataset concepts}
\noindent \textbf{Arts and crafts}: Image must contain arts and crafts.

\noindent \textbf{Astronaut}: Any picture that shows an astronaut, even if it's a drawing, clip art, etc. The astronaut should show clearly that they are associated with being an astronaut – usually indicated by a space suit or NASA jumpsuit.

\noindent \textbf{Block tower}: Image must contain a toy block tower made of legos or wood. 

\noindent \textbf{Dance}: Photos of people dancing.

\noindent \textbf{Emergency service}: Image must contain emergency service,  paramedics, firefighters, police, or rescue teams.

\noindent \textbf{Gourmet tuna}: Photos of gourmet dishes (i.e. fancy, elegant) that must contain tuna. This includes sushi, sashimi, seared tuna, a fancy ahi tuna salad. This does not include canned tuna, tuna sandwich, a photo of the fish tuna itself.

\noindent \textbf{Hand pointing}: A picture showing a hand pointing, with just the index finger extended. Does not include pictures of thumbs-up or pictures of hands with more than just the index finger extended. Picture with a straight finger pointing at or tapping a screen are included.

\noindent \textbf{Hair coloring}: Pictures that focus on people during the process of hair coloring or right after, before \& after photos. Negatives: store front of hairdresser, boxes of dye.

\noindent \textbf{Healthy dish}: Photos of dishes with healthy food that is low in carbs

\noindent \textbf{Home fragrance}: Photos of any types of fragrances used for houses, including home perfumes, air fresheners for the house, scented candles, essential oils. 

\noindent \textbf{In ear headphones}: Any headphones that are worn inside the ear, rather than covering it up. These types of headphones are inserted into the ear canal. As long as an in-ear headphone is in the picture, it is valid.

\noindent \textbf{Pie chart}: Any image with a pie chart, which is a circular statistical graphic, which is divided into slices to illustrate numerical proportion.

\noindent \textbf{Single sneaker on white background}: Images depicting a single sneaker on a white or neutral-colored background (e.g beige). It can be a partial view of a sneaker (e.g. just the sole, or half of the sneaker is in view) but it cannot be just parts (e.g. just the shoe lace) . Negatives include images that have more than one shoe, that have different colored background, or a different style of shoe. 

\noindent \textbf{Stop sign}: This includes photos of real-world, official stop signs. Imagine we would want to detect such stop signs for self-driving cars. Positives include any stop sign photos, including those temporary ones included in construction, or held in hand by a construction worker. If there’s a stop sign on a banner or ads poster, even if it’s in traffic, it would be a negative (we don't want the self-driving car to stop at that). Clip art or indoors stop sign are negative
\subsection{Public dataset concepts}
\noindent \textbf{Hateful memes}: Memes that are harmful, racist, or sexist

\section{Search queries}
The following is the set of search queries used to mine candidate images from the LAION~\cite{schuhmann2022laion} dataset during the data mining process of our system. All these search queries are generated using the LLM (PaLM-2 \cite{anil2023palm2}) and encoded in joint CLIP \cite{radford2021learning} embedding space to retrieve candidate images.

\noindent \textbf{Arts and crafts}:  craft room, crafts book, crafts for beginners, crafts tutorial, arts and crafts, craft store, crafts fair, craft project, crafts for sale, crafts, crafts for kids, art, craftsmanship, craft supplies, diy, crafts magazine, crafts for adults, handmade 

\noindent \textbf{Astronaut}:  astronaut, astronaut in space station module, astronaut in space gloves, astronaut in space boots, astronaut in space flag, astronaut in space shuttle cockpit, astronaut in space suit, astronaut in orbit, astronaut in space backpack, astronaut in space station airlock, astronaut on moon, astronaut working in space, astronaut in space station, astronaut in space, astronaut in space helmet, astronaut in space shuttle, astronaut in space station cupola, astronaut walking in space 

\noindent \textbf{Block tower}:  tower of blocks, lego tower, tall lego tower, tall toy tower, tower of legos, tower of toys, towering wood, towering toys, towering blocks, block tower, towering legos, tall block tower, tower made of blocks, tall wooden tower, wooden tower, toy tower, tower of wood 

\noindent \textbf{Dance}:  street dance, flamenco, ballet, modern dance, bachata, ballroom dance, zouk, samba, people dancing, belly dance, salsa, merengue, dance performance, line dance, tap dance, hip hop, dancers, folk dance 

\noindent \textbf{Emergency service}:  police emergency, police officer at work, rescue boat, emergency response, police car, fire truck, rescue worker at work, emergency worker, medical emergency, firefighter, paramedic at work, rescue team, fire rescue, police, emergency service, rescue operation, firefighter at work, rescue helicopter, emergency vehicle, ambulance, paramedic 

\noindent \textbf{Gourmet tuna}:  tuna sushi, tuna sashimi, tuna salad, seared tuna, ahi tuna, gourmet tuna, tuna tartare, ahi tuna steak, tuna steak, fancy tuna 

\noindent \textbf{Hand pointing}:  hand pointing finger extended, hand pointing finger, hand pointing, hand pointing finger straight at them, hand pointing finger straight at someone, hand pointing finger straight at screen, hand index finger extended, hand pointing finger straight at something, hand pointing finger straight at person, hand pointing finger straight at me, hand pointing finger straight at us, hand pointing finger straight at you, hand pointing at screen, hand pointing finger straight at thing, hand pointing screen, hand pointing finger straight at object, hand pointing finger straight 

\noindent \textbf{Hair coloring}:  hair coloring, hair color salon, hair color before and after, hair color inspiration, hair color horror story, hair color mishap, hair color tutorial, hair color tips, hair dye, hair color stylist, hair color fail, hair color at home, hair color process, hair color mistake, hair color disaster, hair color gone wrong, hair color ideas, hair color, hair color gone bad 

\noindent \textbf{Healthy dish}:  healthy lunch, healthy sandwich, healthy dish, healthy burger, healthy meal, healthy food, low carb dish, healthy salad, healthy fish, healthy pizza, healthy dinner, healthy vegetarian, healthy vegan, healthy snack, healthy breakfast, healthy pasta, healthy chicken, healthy soup, healthy dessert 

\noindent \textbf{Home fragrance}:  home fragrance, home fragrance diffuser, scented candle, home scented candle, home scent, essential oil, home air freshener, air freshener, home essential oil, home scent diffuser, home room spray, home aroma diffuser, home smell diffuser, home perfume, home aroma, fragrance diffuser, home smell, room spray 

\noindent \textbf{In ear headphones}:  earphone, in ear headphones, in ear headphone, earbuds, in ear, headphone 

\noindent \textbf{Pie chart}:  pie chart maker, pie chart data, pie chart tutorial, pie chart percentage, pie chart illustration, pie chart design, pie chart template, pie chart chart, pie chart, pie chart infographic, pie chart graphic, pie chart diagram, pie chart creator, pie chart graph, pie chart example, pie chart generator, pie chart tool, pie chart software 

\noindent \textbf{Single sneaker on white background}:  sneaker on light beige background, sneaker on beige background, sneaker on neutral, sneaker on light background, sneaker on cream background, single sneaker, sneaker on white background, sneaker, sneaker on background, sneaker on solid background, sneaker on light off-white background, sneaker on light tan background, sneaker on neutral background, sneaker on light gray background, sneaker on light cream background, sneaker on plain background, sneaker on off-white background, shoe, sneaker on tan background, sneaker on white, sneaker on gray background 

\noindent \textbf{Stop sign}:  stop sign on road, stop sign on street, held stop sign, stop sign held by person, traffic stop sign, stop sign in traffic, stop sign in city, stop sign on interstate, stop sign, stop sign on highway, construction stop sign, stop sign in construction, real stop sign, stop sign on freeway, stop sign in rural area, official stop sign, stop sign in parking lot, stop sign in hand 

\section{LLM Prompts}
We list a set of example prompts used in Modeling Collaborator Annotator below. When a description is unavailable for a given concept, we use the following prompt to auto-generate a structured description:
\begin{lstlisting}
You are given a visual concept name.

Follow these steps:
<step1>You have to work as an expert linguist. There are some human annotators who need to determine if given images are in-scope or out-of-scope for this visual concept. Your task is to generate description of the visual concept which annotators can use to decide if any images are in-scope or out-of-scope for the given visual concept.</step1>
<step2>Provide an concept definition of this visual image in a few sentences.</step2>
<step3>Provide all the image attributes that an image must have in order to be in-scope for this visual concept.</step3>
<step4>Each attribute found in step2 and step3 should be verbose, independent, self-explanatory and meaningful.</step4>
<step7>Write your response in following user friendly and readable format:
Visual concept definition:
<Add 2-3 line concept definition of the visual concept here.>

Image must have following attributes for it to be in-scope for this visual concept:
<Add details here as bullet points.>
</step7>
<visualConceptName>{CONCEPT_NAME}</visualConceptName>
\end{lstlisting}

The prompt for generating positive search queries based on a visual concept (used to fetch candidate positive images from an image database):
\begin{lstlisting}
Your task is to help in finding positive (in-scope) images for a visual concept. You are given the name and the description of a visual concept. Description explains the attributes of an image that make it in-scope for this visual concept. It also explains the attributes of an image that make it out-of-scope for this visual concept.

Follow these steps:
<step1>List all the attributes of an image that make it in-scope for this visual concept.</step1>
<step2>Each attribute should be objective, complete, and self-explanatory.</step2>
<step3>Ensure that attributes you have found cover all the in-scope attributes or scenarios mentioned in the description. If not, add the missing in-scope attributes.</step3>
<step4>Based on all the in-scope attributes you have identified in step3, generate 20 Google Search keywords which can be used to do Google image search for finding diverse images with those in-scope attributes.</step4>
<step5>Ensure that your Google Search keywords cover all types of in-scope images mentioned in the description. If not, add Google Search keywords to find those  types of in-scope images.</step5>
<step6>This is an important step. Some of the keywords you selected so far could be be suitable for searching out-of-scope images. Identify those keywords which are for searching out-of-scope image. Remove those Google Search keywords from your response.</step6>
<step7>Each search query should be 3-4 words long, independent, self-explanatory and meaningful for internet image search.</step7>
<step8>If any of these queries are longer than 4 words, summarize them into 3-4 words.</step8>

<step9>Write your response in following xml format. Since your response will be programmatically parsed, your response should strictly follow this format:
```xml
<google_search_keywords>
 <keyword></keyword>
 ...
</google_search_keywords>
```
</step9>
<step10>Keep only xml in the response and remove other text.</step10>
<concept>{CONCEPT_NAME}</concept>
<description>{CONCEPT_DESCRIPTION}</description>
\end{lstlisting}

The prompt for generating negative search queries based on a visual concept  (used to fetch hard negative images from an image database):
\begin{lstlisting}
You have to work as an expert linguist.  You are given a visual concept name and its description for the purpose of image classification.
Description might contains few carve-outs. Carve-outs are some special situations in which images should be classified as out-of-scope. Your task is to extract carve-out details from the description.

Follow these steps:
<step1>If the description does not contain any carve-outs, write your response in the following format and skip all of the following steps.
```xml
<carveOutsInDescription>
  <carveOut>NOT_FOUND</carveOut>
</carveOutsInDescription>
```
</step1>
<step2>If the description provides details on out-of-scope images for this visual concept, output the list of those carve-outs situations mentioned.</step2>
<step3>Output those in the following xml format. Since your response will be programmatically parsed, your response should strictly follow this format:
```xml
<carveOutsInDescription>
  <carveOut></carveOut>
  ...
</carveOutsInDescription>
```
</step3>
<step4>Keep only xml in the response and remove other text.</step4>

<concept>{CONCEPT_NAME}</concept>
<description>{CONCEPT_DESCRIPTION}</description>

\end{lstlisting}

The prompt template for generating a concept's positive attributes:

\begin{lstlisting}
Your task is to understand the scope of a visual concept for image classification. You are given a visual concept name and its description.

Description explains the attributes of an image that make it in-scope for this visual concept. It also explains the attributes of an image that make it out-of-scope for this visual concept.

Follow these steps:

<step1>List all the attributes of an image that make it in-scope for this visual concept.</step1>

<step2>Each attribute should be objective, unambiguous, detailed, verbose and self-explanatory.</step2>

<step3>Check that attributes you have found cover all the positive attributes mentioned in the description. If not, add the missing attributes.</step3>

<step4>Write your response in following xml format. Since your  response will be programmatically parsed, your response should strictly follow this format:

```xml
<positiveAttributes>
  <attribute></attribute>
  ...
</positiveAttributes>
```
</step4>
<step5>Keep only xml in the response and remove other text.</step5>
<concept>{CONCEPT_NAME}</concept>
<description>{CONCEPT_DESCRIPTION}</description>
\end{lstlisting}
The prompt template for generating a concept's negative attributes:
\begin{lstlisting}
You have to work as an expert linguist.  You are given a visual concept name and its description for the purpose of image classification.
Description might contains few carve-outs. Carve-outs are some special situations in which images should be classified as out-of-scope. Your task is to extract carve-out details from the description.

Follow these steps:
<step1>If the description does not contain any carve-outs, write your response in the following format and skip all of the following steps.
```xml
<carveOutsInDescription>
  <carveOut>NOT_FOUND</carveOut>
</carveOutsInDescription>
```
</step1>
<step2>If the description provides details on out-of-scope images for this visual concept, output the list of those carve-outs situations mentioned.</step2>
<step3>Output those in the following xml format. Since your response will be programmatically parsed, your response should strictly follow this format:
```xml
<carveOutsInDescription>
  <carveOut></carveOut>
  ...
</carveOutsInDescription>
```
</step3>
<step4>Keep only xml in the response and remove other text.</step4>

<concept>{CONCEPT_NAME}</concept>
<description>{CONCEPT_DESCRIPTION}</description>
\end{lstlisting}
where \texttt{CONCEPT\_NAME} and \texttt{CONCEPT\_DESCRIPTION} are the subjective concept name and description. 

For a final annotation decision for an image, we feed the following prompt to PaLM-2:
\begin{lstlisting}
You are given the name and description of a visual concept. We showed the image to raters and asked many questions about the image and they gave the  answers. Questions and answers are also provided below. Your task is to answer some questions about the image. Follow these steps:

<step1>In the following steps, your text responses must be strictly based on the answers provided in raters' responses.</step1>

<step2>Provide the out-of-scope attributes present in the image.</step2>

<step3>Provide the in-scope attributes present in the image.</step3>

<step4>Provide the the in-scope attributes missing in the image.</step4>

<step5>Classify the image based on the following rules. Rules must be followed in the given order.

<classificationRules>

<rule1>If  the image has EVEN ONE of the out-of-scope attributes, it must be classified negative for this visual concept.</rule1>

<rule2>The image must have all the required positive attributes to classify the image as positive for this visual concept.   If image has all the required positive attributes, classify the image as positive. Otherwise classify it as negative.</rule2>

<rule3>In all other cases, classify the image as negative.</rule3>
</classificationRules></step5>

<step6>Add following details to your response strictly in this format:
Decision: "Positive" or "Negative"
Reasons: <Provide list of reasons why this image is Positive or Negative> </step6>

<step7>Make sure your response only contains text and no python code.</step7>

<concept>{CONCEPT_NAME}</concept>
<conceptDescription>{CONCEPT_DESCRIPTION}</conceptDescription>
<raterResponses>{PALI_QUESTIONS_AND_ANSWERS}</raterResponses>
\end{lstlisting}
where \texttt{PALI\_QUESTIONS\_AND\_ANSWERS} is a formatted string of questions fed to PaLI-X VQA and their respective answers.

\section{Ablations}
\begin{figure*}[]
    \centering
    \includegraphics[width=\linewidth]{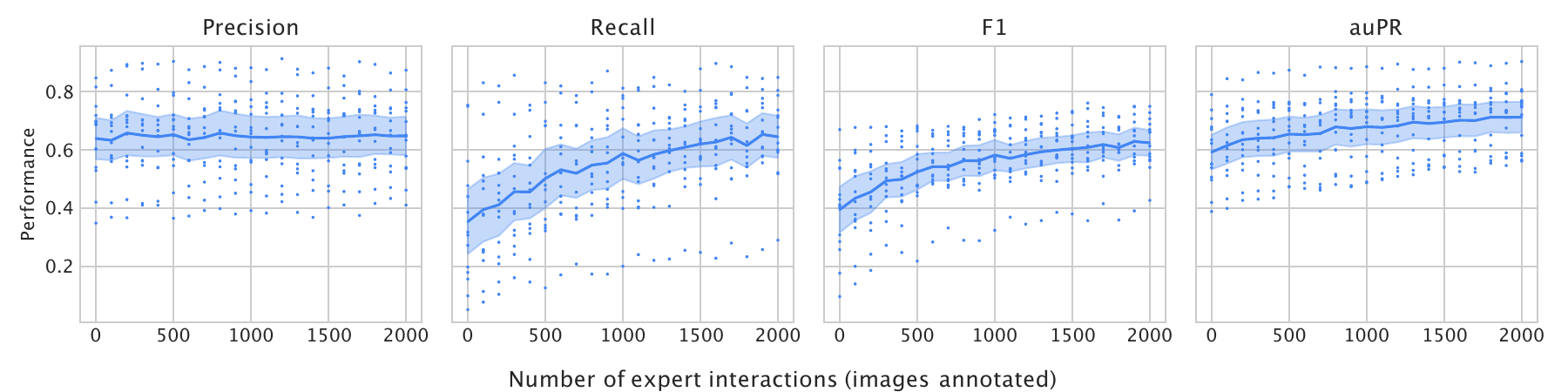}
    \caption{The impact of Modeling Collaborator and expert collaboration on the performance of the distilled model. 4,000 total training examples were used per concept. The x-axis represents how many of those examples were labeled by the expert (concept owner), ranging from no examples (0\%) to 2,000 examples (50\%).}
    \label{fig:results-expert-annotations-impact}
\end{figure*}

To measure the effect of the expert involvement we show \cref{fig:results-expert-annotations-impact}. Overall, expert collaboration improves the performance of the distilled model. As the number of expert-labeled examples increases (0 to 2000 out of total training 4000 examples), the recall, F1, and auPR scores of the model also increase.

\begin{figure}[ht]
    \centering
    \includegraphics[width=0.9\linewidth]{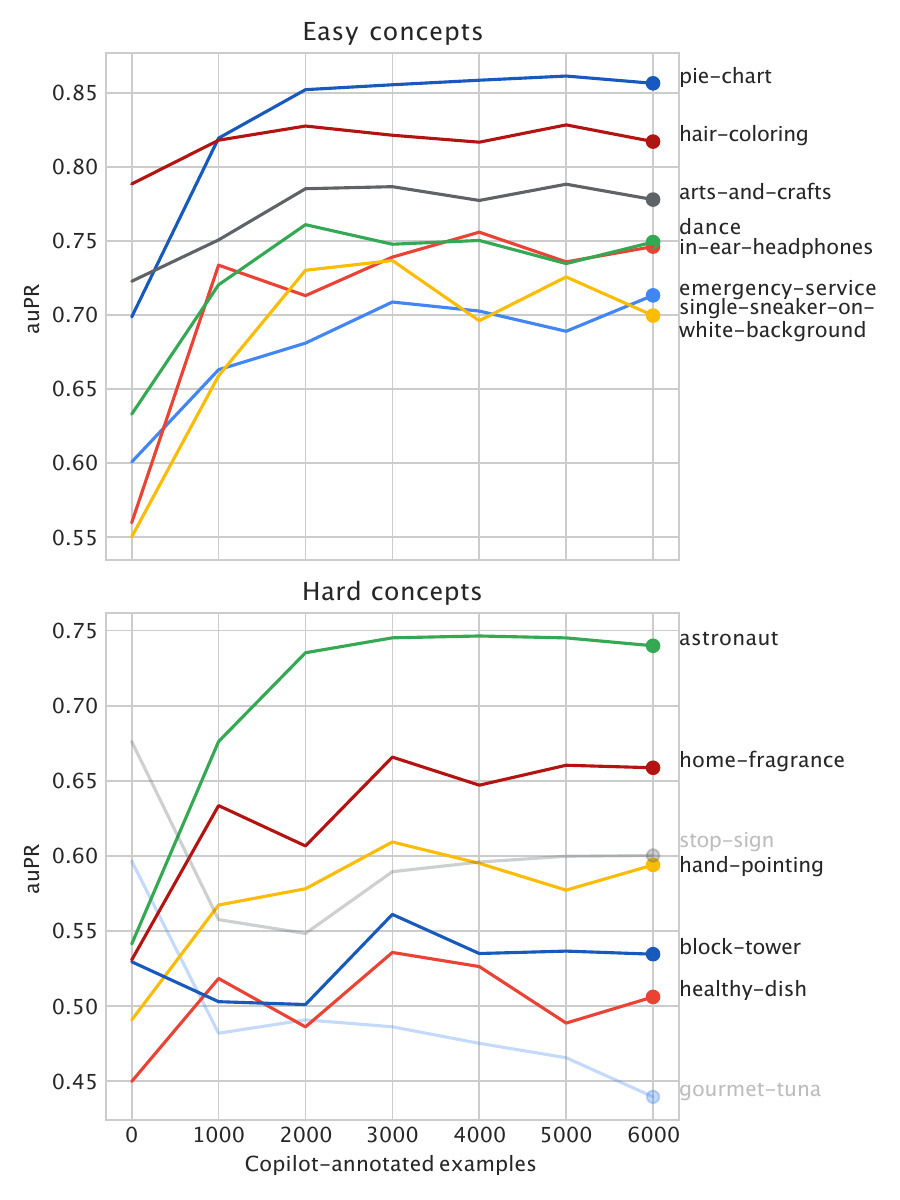}
    \caption{The impact of adding additional automatically annotated images on the final model quality (using the auPR metric). 100 user-annotated examples are used in addition to the thousands of Modeling Collaborator examples.}
    \label{fig:results-modeling-copilot-per-concept}
\end{figure}

To show the impact of additional automatically annotated data on the performance of the final output model on easy vs hard concepts, we show \cref{fig:results-modeling-copilot-per-concept}.
\section{Annotator Configurations}
We define the following settings for the Annotator:
\begin{enumerate}[label=\Alph*.]
    \item \texttt{use\_positive\_attributes\_for\_questions}: Whether to generate positive questions from the attributes or directly from the concept description.
    
    \item  \texttt{generate\_negative\_questions}: Whether or not to generate negative questions. Sometimes these questions result in over-predicting negative classes.
    
    \item  \texttt{use\_captioning\_questions}: Whether to use a captioning VLM to generate a detailed description of the image
    
    \item \texttt{generate\_fixed\_num\_of\_questions}: Fix the number of questions instead of having the LLM generate as many questions as possible.
    
    \item  \texttt{final\_rating\_without\_attributes}: Whether to use negative and positive attribute in the final annotation stage. 
\end{enumerate}

Using a grid search, we use different configurations for different concept as described in \cref{tab:configs}.

\begin{table}[]
\centering
\begin{tabular}{|l|ccccc|}
\hline
                                   & \multicolumn{5}{c|}{Configuration}                             \\ \hline
\textbf{Concept}                   & \textbf{A} & \textbf{B} & \textbf{C} & \textbf{D} & \textbf{E} \\ \hline
arts-and-crafts                    &            &            &            & \checkmark & \checkmark \\
astronaut                          &            &            & \checkmark & \checkmark & \checkmark \\
block-tower                        &            &            & \checkmark & \checkmark & \checkmark \\
dance                              &            &            & \checkmark & \checkmark & \checkmark \\
emergency-service                  &            &            &            &            & \checkmark \\
gourmet-tuna                       &            &            &            & \checkmark & \checkmark \\
hair-coloring                      &            &            &            &            & \checkmark \\
hand-pointing                      &            &            &            &            & \checkmark \\
healthy-dish                       &            &            & \checkmark & \checkmark & \checkmark \\
home-fragrance                     &            &            &            & \checkmark & \checkmark \\
in-ear-headphones                  &            &            & \checkmark & \checkmark & \checkmark \\
pie-chart                          &            &            & \checkmark & \checkmark & \checkmark \\
single-sneaker                     &            &            &            & \checkmark & \checkmark \\
stop-sign                          &            &            &            & \checkmark & \checkmark \\ \hline
\end{tabular}
\caption{Configuration settings used for each concept of the Agile Modeling \cite{stretcu2023agile} dataset.}
\label{tab:configs}
\end{table}

\end{appendix}

\end{document}